\pdfoutput=1

\documentclass[11pt]{article}

\usepackage[]{latex/acl}

\usepackage{times}
\usepackage{latexsym}

\usepackage[T1]{fontenc}

\usepackage[utf8]{inputenc}

\usepackage{microtype}

\usepackage{inconsolata}

\usepackage[utf8]{inputenc}
\usepackage{booktabs}
\usepackage{graphicx}
\usepackage{CJK}
\usepackage{multirow}
\usepackage{color}
\usepackage{verbatim}
\usepackage{url}
\usepackage{enumitem}
\usepackage{amsmath}
\usepackage{diagbox}
\usepackage{flushend,cuted}
\usepackage{bbm}
\usepackage{xspace}
\usepackage{algorithm}
\usepackage{algpseudocode}
\usepackage{colortbl}
\usepackage{bbding}
\usepackage{amssymb}
\DeclareMathOperator*{\argmax}{arg\,max}

\usepackage{listings}
\usepackage{ulem}
\usepackage{listings}
\usepackage{wrapfig}
\usepackage{soul}
\usepackage{makecell}

\usepackage{xcolor}

\newcommand{\kaiti}[1]{\begin{CJK*}{UTF8}{gkai} #1 \end{CJK*}}
\soulregister{\kaiti}7
\usepackage{pifont}
\usepackage{arydshln}
\usepackage{booktabs}

\title{Question Translation Training for Better Multilingual Reasoning}

\author{
    Wenhao Zhu$^{1}$\text{,} \textbf{Shujian Huang}$^{1}$\text{,} \textbf{Fei Yuan}$^{2}$\text{,} \textbf{Shuaijie She}$^{1}$\text{,} \textbf{Jiajun Chen}$^{1}$\textbf{,} \textbf{Alexandra Birch}$^{3}$ \\
    $^{1}$ \text{National Key Laboratory for Novel Software Technology, Nanjing University} \\
    $^{2}$ \text{Shanghai AI Lab}  $^{3}$ \text{School of Informatics, University of Edinburgh} \\
    \small\texttt{zhuwh@smail.nju.edu.cn}, \small\texttt{huangsj@nju.edu.cn}, \small\texttt{yuanfei@pjlab.org.cn}, \small\texttt{shesj@smail.nju.edu.cn} \\
    \small\texttt{chenjj@nju.edu.cn}, \small\texttt{a.birch@ed.ac.uk} \\
}

\begin{document}
\maketitle

\begin{abstract}
Large language models show compelling performance on reasoning tasks but they tend to perform much worse in languages other than English. 
This is unsurprising given that their training data largely consists of English text and instructions. 
A typical solution is to translate  instruction data into all languages of interest, and then train on the resulting multilingual data, which is called translate-training. This approach not only incurs high cost, but also results in poorly translated data due to the non-standard formatting of mathematical chain-of-thought.
In this paper, we explore the benefits of question alignment, where we train the model to translate reasoning questions into English by finetuning on X-English parallel question data. 
In this way we perform targeted, in-domain language alignment which makes best use of English instruction data to unlock the LLMs' multilingual reasoning abilities.
Experimental results on LLaMA2-13B show that question alignment leads to consistent improvements over the translate-training approach: an average improvement of 11.3\% and 16.1\% accuracy across ten languages on the MGSM and MSVAMP multilingual reasoning benchmarks\footnote{The project will be available at: \url{https://github.com/NJUNLP/QAlign}.}.
\end{abstract}

\section{Introduction}
Large language models have recently shown a strong ability to reason in English, but performance in other languages, especially more distant languages, still trails far behind~\cite{shi2022language,huang2023not}.
It is unsurprising, considering that their training data is predominantly composed of English text and instructions~\cite{blevins2022language,touvron2023llama,wang2023far}.
To elicit LLM's multilingual performance, previous approach typically follows the translate-training paradigm~\cite{chen2023breaking}, which first translates English instruction data into non-English with a translation engine and then uses the multilingual data for instruction-tuning.

However, the translate-training has the following drawbacks: (1) translating English training data to numerous non-English languages incurs significant translation cost, especially considering the constant addition of large and complex instruction tuning sets~\cite{yuan2023scaling,yu2023metamath}.
(2) Additionaly, it is hard for the translation engine to accurately translate lengthy, logical texts containing mathematical symbols in chain-of-thought (CoT) responses, which can compromise the quality of translated data (evidence are shown in Appendix~\ref{sec:quality}).
Consequently, we explore the following research question in this paper: \textit{Can we unlock the LLM's multilingual reasoning ability by teaching it to translate reasoning questions into English?}

In this paper, we focus on the multilingual mathematical reasoning task and explore the benefits of question alignment (QAlign), where we fine-tune the pre-trained LLM to translate reasoning questions into English with X-English parallel question data.
This targeted, in-domain language alignment enables the subsequent effective utilization of English instruction data to unlock LLMs' multilingual reasoning abilities.
Following question alignment, we implement response alignment by further fine-tuning the language-aligned LLM with cutting-edge English instruction data.
Even when only English supervised data is available, our alignment-enhanced LLM can achieve superior performance on non-English tasks with its transferable English expertise.

To demonstrate the advantages of question alignment, we conduct experiments on challenging multilingual mathematical reasoning benchmarks, \textsc{mGSM}~\cite{shi2022language} and \textsc{mSVAMP}~\cite{chen2023breaking}.
We use two of the most advanced open-source LLMs, LLaMA2-7B and LLaMA2-13B~\cite{touvron2023llama}, as base models.
Experiment results show that the inclusion of the question alignment stage brings an average improvement of up to 13.2\% in multilingual performance.
The performance improvement on low-recourse languages, e.g. Thai and Swahili, can be 30\%-40\%.
Compared to the translate-training baseline, MathOctopus ~\cite{chen2023breaking}, which tuned with a multilingual version of \textsc{GSM8K} dataset , our alignment-enhanced LLMs achieves average performance improvement of 9.6\%~(7B) and 11.3\% (13B) on \textsc{mGSM}.
On the out-of-domain test set \textsc{mSVAMP}, our fine-tuned LLMs achieve 13.1\% (7B) and 16.1\% (13B) average accuracy improvement, also demonstrating our approach is robust to domain shift.
In general, we observe that although incorporating translated instruction data does benefit multilingual performance, our question alignment strategy provides a more efficient and effective choice.
In our analysis, we also present the effects of other implementations for performing language alignment and illustrate the importance of choosing the appropriate translation direction and domain during this phase of training.

The main contributions of this paper can be summarized as:
\begin{itemize}[itemsep=1pt]
    \item We present a novel X-English question alignment finetuning step which performs targeted language alignment for best use of the LLMs English reasoning abilities.
    \item We fine-tune open-source LLMs, LLaMA2-7B/13B, into strong multilingual reasoners, which beat the translate-training baseline by 9.6\% (7B) and 11.3\% (13B) on \textsc{mGSM}, by 13.1\% (7B) and 16.1\% (13B) on \textsc{mSVAMP}.
    \item We explore language alignment with other language directions (English-X), types and domains of data, and confirm our intuition that in fact X-English questions perform best. 
\end{itemize}

\section{Related Work}
\noindent\paragraph{Large language model}
With a large number of parameters pre-trained on a large-scale corpora, large language models can memorize vast amounts of knowledge~\cite{roberts2020much} and acquire emergent abilitie, such as in-context learning~\cite{brown2020language}, CoT generation~\cite{wei2022chain}.
Then, to better align the behavior of LLMs with human expectations, \citet{wei2022finetuned} propose instruction-tuning, training LLM to generate desired response based on the given instruction.
Subsequently, many efforts are put into creating effective instruction data to further unlock LLM's potential~\cite{wang2022super,alpaca,longpre2023flan,wang2023far}.
However, since the proposed instruction datasets consist mainly of English, the directly fine-tuned LLMs struggle on non-English languages, especially on those languages that are dissimilar to English~\cite{huang2023not,zhu2023extrapolating,chen2023breaking}.

\noindent\paragraph{Multilingual mathematical reasoning}
Mathematical reasoning is a challenging and representative task for evaluating the intelligence of LLMs~\cite{ahn2024large}, where LLMs need to understand the given math question and produce a numerical answer through step-by-step reasoning.
\citet{shi2022language} expanded the scope to a multilingual context by translating English math questions from the \textsc{GSM8K} test set~\cite{cobbe2021training} into non-English languages, thereby creating a multilingual benchmark called \textsc{mGSM}.

Subsequently, many efforts are put into enhancing LLM's multilingual reasoning capabilities, which can be categorized into two approaches: prompting close-source LLMs and instruction-tuning open-source LLMs.
In the first approach, \citet{qin2023cross} and \citet{huang2023not} carefully craft prompts for close-source LLMs like ChatGPT~\cite{openai2022chatgpt}.
Their strategy involves first prompting the LLM to explicitly translate non-English questions into English, then ask the model to solve the translated problem instead.
However, the effectiveness of these prompting methods are not well-examined on open-source LLMs.
And it remains an open challenge to equip open-sourced LLMs with strong multilingual mathematical problem-solving skills.

In the second approach, \citet{chen2023breaking} follow the translate-training method~\cite{artetxe2023revisiting}.
Initially, they translate English instruction data in \textsc{GSM8K} into non-English with ChatGPT, followed by employing multilingual data for instruction-tuning.
\begin{figure*}
    \centering
    \includegraphics[width=1\textwidth]{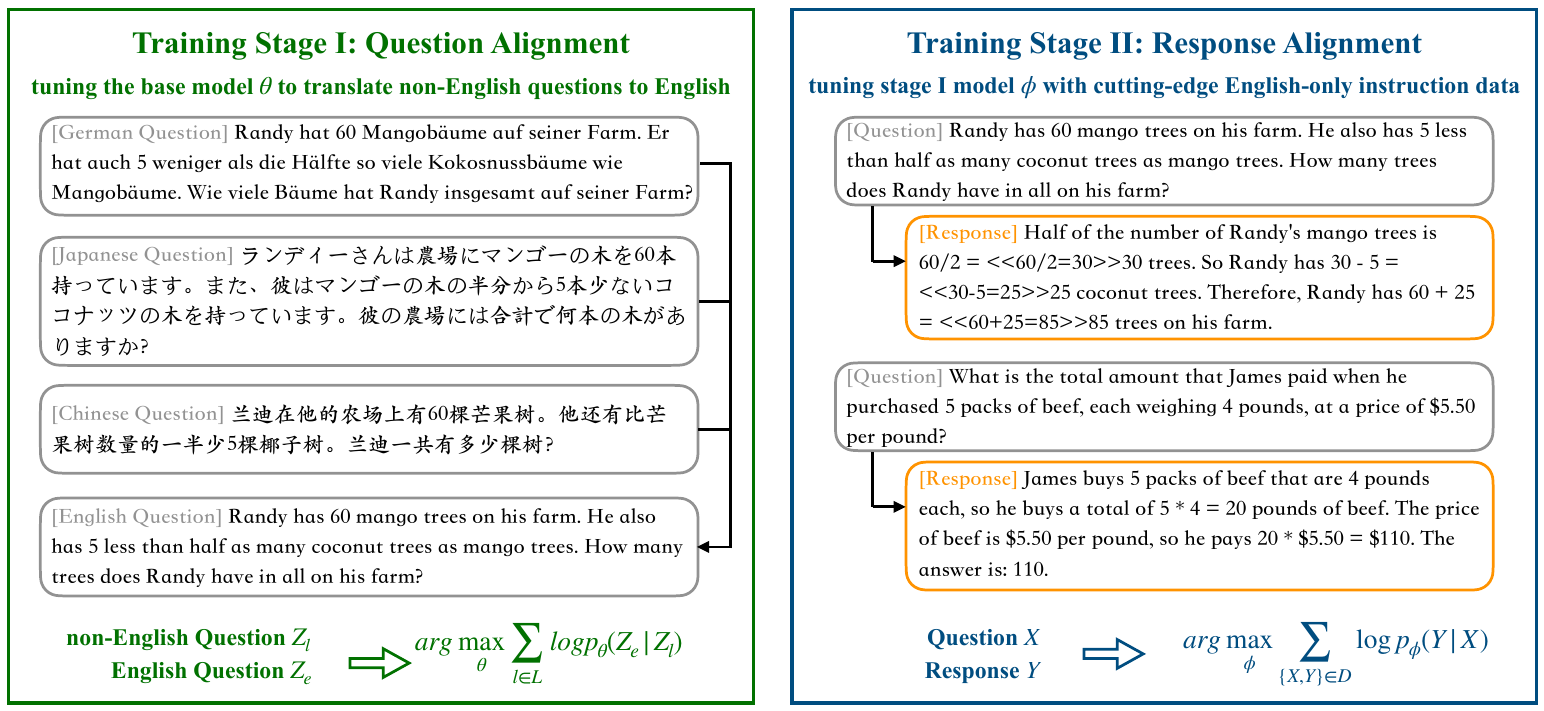}
   \caption{Illustration of our devised two-step training framework. At training stage I (question alignment), we use a set of multilingual questions for translation training. At training stage II (response alignment), we use cutting-edge English-only instruction data for fine-tuning. Due to the established language alignment in stage I, we can utilize LLM's expertise in English to enhance its performance on non-English tasks.}
   \label{fig:illustration}
\end{figure*}
Moreover, \citet{chen2023breaking} investigate cross-lingual training strategies such as mixing questions and CoT responses in different languages, but fail to achieve consistent improvement.
Although the translate-training approach is effective, it incurs high translation cost and is error-prone\footnote{We analyze the errors in the translated dataset from ~\citet{chen2023breaking} and present both quantitative and qualitative results in Appendix~\ref{sec:quality}.}.
It also becomes increasingly impractical to translate vast quantities of augmented data into numerous languages, especially considering recent findings that augmented training data, e.g., \textsc{MetaMathQA}~\cite{yu2023metamath}—which is 50 times larger than \textsc{GSM8K}—greatly enhances LLM's reasoning skills. 
Without relying on translated CoT responses, in this paper, we present a novel question alignment technique to utilize cutting-edge English-only supervised data to boost open-source LLM's performance on multilingual reasoning tasks.

\section{Methodology}
An illustration of our devised method is shown in Figure~\ref{fig:illustration}.
The key idea of our approach is strengthening language alignment within LLM before exposing it to English instruction-response pairs.
By doing so, we can utilize LLM's expertise in English to enhance its performance on non-English tasks. 
Below we introduce the two training stages of our framework: question alignment (\S\ref{sec:language_alignment}) and response alignment (\S\ref{sec:instruction_alignment}).

\subsection{Stage I: Question Alignment}
\label{sec:language_alignment}
It has been found that directly fine-tuning LLMs with English instruction data does not help to improve their performance on non-English tasks~\cite{chen2023breaking}.
We suggest that this issue may arise from the insufficient alignment of multiple languages within the LLM. 
Ideally, in a well-aligned LLM, proficiency in one language, like English, could easily transfer to other languages.

To improve the alignment of non-English languages with English, we devise a translation task \textbf{QAlign}: training LLM on translating questions from non-English into English.
Specifically, given a group of multilingual questions, the optimization objective can be written as:
\begin{equation}
\argmax_{\theta} \sum_{l\in\mathcal{L}} \log p_{\theta}(\mathcal{Z}_{e} | \mathcal{Z}_l) \nonumber
\end{equation}
where $\theta$ denotes the parameters of the base model. $\mathcal{Z}_l$ and $\mathcal{Z}_e$ denote non-English and English questions respectively and $\mathcal{L}$ is the set of considered non-English languages.
With this training objective, we equip the LLM with an implicit bias to relate non-English questions with their English counterparts when performing non-English tasks.

Note that this stage only relies on multilingual questions rather than translated CoT responses.
Basically, acquiring multilingual questions is more feasible than obtaining accurate multilingual CoT responses, because translation engines often struggle to precisely translate lengthy, logical texts containing mathematical symbols (quantitative evidence are shown in Appendix~\ref{sec:quality}).

In this translation task, the domain of translation data is also an important factor to consider.
In subsequent experiments, we demonstrate that using multilingual questions as translation data is more effective than employing general domain translation corpora. 

\subsection{Stage II: Response Alignment}
After question alignment, we train LLM with specialized instruction-response pairs to unlock its potential on multilingual mathematical reasoning tasks.
Specifically, we consider two data scenarios: monolingual supervision setting and mixed supervision setting.

\label{sec:instruction_alignment}
\noindent\paragraph{Monolingual supervision setting}
In this setting, we employ English-only instruction data for response alignment, because the cutting-edge instruction datasets are often available only in English.
During training, we follow the standard implementation~\cite{wei2022finetuned} and finetune the language-aligned LLM to maximize the generetive probability of the response $\mathcal{Y}$ given the question $\mathcal{X}$:
\begin{equation}
\argmax_{\phi} \sum_{\{\mathcal{X},\mathcal{Y}\}\in\mathcal{D}} \log p_{\phi}(\mathcal{Y} | \mathcal{X}) \nonumber
\end{equation}
Where $\phi$ denotes the parameters of the stage I model and $\mathcal{D}$ denotes the instruction dataset.
Although the training only utilizes English supervision, the previously established language alignment allows the LLM's English proficiency to be shared across multiple languages.

\noindent\paragraph{Mixed supervision setting}
While our framework is primarily designed for the scenario where only English instruction data is available, it can also leverage additional multilingual supervision, when available, to achieve even higher multilingual performance.
For instance, this multilingual dataset could be a translated version of a subset of large-scale English data.
In this scenario, given a set of additional multilingual superivsed data $\mathcal{M}$, we sequentially fine-tune the stage I model on $\mathcal{M}$ and then on the English instruction data $\mathcal{D}$.
Subsequent experiment results show that this training recipe can further improve the LLM's multilingual reasoning capabilities.

\section{Experiment Setting}
\noindent\paragraph{Base LLM}
In our experiments, we use two of the most advanced open-source LLMs, LLaMA2-7B and LLaMA2-13B as the base model.

\begin{table}[ht]
\centering
\footnotesize
\begin{tabular}{cccc}
\toprule
\textbf{Dataset}       & \textbf{Usage} & \textbf{\# Lang} & \textbf{\# Sample}  \\
\midrule
\textsc{MetaMathQA}    & Training   & 1      & 395,000 \\
\textsc{GSM8KInstruct} & Training   & 10     & 73,559  \\
\midrule
\textsc{mGSM}          & Evaluation & 10     & 2,500   \\
\textsc{mSVAMP}        & Evaluation & 10     & 10,000  \\
\bottomrule
\end{tabular}
\caption{Statistics of involved datasets. ``\# Lang'' denotes the number of languages covered by the dataset and ``\# Sample'' refers to the total number of samples it contains.}
\label{tab:statistics}
\end{table}
\noindent\paragraph{Training Dataset}
In the question alignment stage, we utilize multilingual questions from \textsc{GSM8KInstruct}\footnote{\textsc{GSM8KInstruct} is a multilingual dataset that extends the English instruction dataset \textsc{GSM8K} by translating English instructions and CoT responses into nine non-English languages with ChatGPT.}~\cite{chen2023breaking}.
During the response alignment stage, we employ the cutting-edge English-only dataset \textsc{MetaMathQA} as monolingual supervision, which is built upon English dataset \textsc{GSM8K}~\cite{cobbe2021training} and \textsc{Math}~\cite{hendrycks2021measuring} by performing data augmentation, such as rephrasing questions and enriching answers.
In the mixed supervision setting, we employ both \textsc{MetaMathQA} and \textsc{GSM8KInstruct}.
Dataset statistics are reported in Table~\ref{tab:statistics}.

\noindent\paragraph{Training Details}
We use \textit{stanford\_alpaca}\footnote{\url{https://github.com/tatsu-lab/stanford_alpaca}} as our code base.
We use consistent training hyper-parameters across two stages of training.
At each stage, we fine-tune LLM's full parameters for 3 epoch on eight NVIDIA A100 GPUs.
The learning rate is set to 2e-5, with a batch size of 128.

\begingroup
\renewcommand{\arraystretch}{1.2}
\begin{table*}[ht]
\centering
\footnotesize
\begin{tabular}{l|cccccccccc|c}
\hline
\hspace{1.2cm}\textbf{System (7B)} & \textbf{Bn} & \textbf{Th} & \textbf{Sw} & \textbf{Ja} & \textbf{Zh} & \textbf{De} & \textbf{Fr} & \textbf{Ru} & \textbf{Es} & \textbf{En} & \textbf{Avg.} \\
\hline
SFT$^\dagger$~\cite{touvron2023llama} & 3.2 & 4.8 & 5.2 & 15.2 & 22.4  & 37.2 & 34.4 & 28.0 & 32.4 & 43.2 & 22.6 \\ 
RFT$^\dagger$~\cite{yuan2023scaling} & 2.4  & 2.0  & 2.8  & 6.8  & 16.8 & 33.6 & 34.0 & 29.2 & 34.0 & 44.8 & 20.6 \\
MAmmoTH$^\dagger$~\cite{yue2023mammoth} & 3.6  & 4.8  & 2.4  & 10.8 & 17.2 & 33.2 & 32.8 & 26.0 & 32.4 & 49.6 & 21.3 \\
WizardMath$^\dagger$~\cite{luo2023wizardmath} & 2.0  & 4.0  & 3.4  & 24.0 & 22.4 & 30.4 & 30.4 & 30.8 & 34.8 & 47.6 & 23.0 \\
MathOctopus$^\dagger$~\cite{chen2023breaking} & 28.8 & 34.4 & 39.2 & 36.0 & 38.4 & 44.8 & 43.6 & 39.6 & 42.4 & 52.4 & 40.0 \\ 
MetaMath~\cite{yu2023metamath} & 6.4 & 4.0 & 3.2 & 39.2 & 38.8 & 56.8 & 52.8 & 47.2 & 58.0 & 63.2 & 37.0 \\
\hdashline
MultiReason & 26.8 & 36.0 & 36.8 & 33.2 & 42.4 & 42.8 & 40.8 & 42.4 & 42.8 & 47.2 & 39.1 \\ 
MonoReason & 7.6  & 5.6  & 5.2  & 34.0 & 45.2 & 54.0 & \textbf{56.8} & 51.6 & 58.8 & 65.5 & 38.4 \\ 
QAlign$\rightarrow$MonoReason (Ours) & \textbf{32.4} & \textbf{39.6} & \textbf{40.4} & \textbf{44.0} & \textbf{48.4} & \textbf{54.8} & \textbf{56.8} & \textbf{52.4} & \textbf{59.6} & \textbf{68.0} & \textbf{49.6} \\
\hline
\hline
\hspace{1.2cm}\textbf{System (13B)} & \textbf{Bn} & \textbf{Th} & \textbf{Sw} & \textbf{Ja} & \textbf{Zh} & \textbf{De} & \textbf{Fr} & \textbf{Ru} & \textbf{Es} & \textbf{En} & \textbf{Avg.}   \\
\hline
SFT$^\dagger$~\cite{touvron2023llama} & 6.0 & 6.8 & 7.6 & 25.2 & 32.8 & 42.8 & 40.8 & 39.2 & 45.2 & 50.4 & 29.7 \\ 
RFT$^\dagger$~\cite{yuan2023scaling} & 3.2 & 4.4 & 3.6 & 26.4 & 33.6 & 38.4 & 44.8 & 41.6 & 46.8 & 52.0 & 29.5 \\
MAmmoTH$^\dagger$~\cite{yue2023mammoth} & 3.6 & 5.2 & 1.6 & 19.2 & 31.2 & 45.6 & 39.6 & 36.8 & 50.0 & 56.4 & 28.9 \\
WizardMath$^\dagger$~\cite{luo2023wizardmath} & 6.4 & 5.6 & 5.6 & 22.0 & 28.0 & 40.4 & 42.0 & 34.4 & 45.6 & 52.8 & 28.3 \\
MathOctopus$^\dagger$~\cite{chen2023breaking} & 35.2 & 46.8 & 42.8 & 43.2 & 48.8 & 44.4 & 48.4 & 47.6 & 48.0 & 53.2 & 45.8 \\ 
MetaMath~\cite{yu2023metamath} & 11.6 & 6.4 & 7.6 & 42.8 & 49.2 & \textbf{64.8} & \textbf{65.2} & 63.6 & 65.2 & 67.2 & 44.4 \\ 
\hdashline
MultiReason & 37.6 & 42.2 & 44.0 & 43.2 & 53.6 & 47.6 & 54.0 & 48.0 & 54.8 & 56.4 & 48.1 \\ 
MonoReason & 12.4 & 11.2 & 6.4 & 42.0 & 46.0 & 64.0 & 62.4 & 61.6 & 64.8 & 68.4 & 43.9 \\
QAlign$\rightarrow$MonoReason (Ours) & \textbf{38.4} & \textbf{49.6} & \textbf{46.0} & \textbf{52.4} & \textbf{59.2} & 62.0 & 62.4 & \textbf{64.4} & \textbf{67.2} & \textbf{69.2} & \textbf{57.1} \\
\hline
\end{tabular}
\caption{Results on \textsc{mGSM} dataset. ``Avg.'' represents the average multilingual performance and bold text denotes the highest score among systems of the same size. The dagger symbol denotes that the results for these models are taken from the published results of~\citet{chen2023breaking}.}
\label{tab:mgsm}
\end{table*}
\endgroup
\noindent\paragraph{Baseline Systems} For comparison, we consider following systems which are instruction-tuned from LLaMA2 with diverse training recipes:
\begin{itemize}[itemsep=0.5pt]
    \item \textbf{SFT}~\cite{touvron2023llama}, which is instruction-tuned with basic \textsc{GSM8K}.
    \item \textbf{RFT}~\cite{yuan2023scaling}, which is instruction-tuned with an augmented \textsc{GSM8K} training dataset, using rejection sampling techniques.
    \item \textbf{MAmmoTH}~\cite{yue2023mammoth}, which is instruction-tuned with \textsc{GSM8K} and a collection of math instruction datasets.
    \item \textbf{WizardMath}~\cite{luo2023wizardmath}, which is constructed using reinforcement learning on \textsc{GSM8K} and \textsc{Math}.
    \item \textbf{MathOctopus}~\cite{chen2023breaking}, which is instruction-tuned with a multilingual version of \textsc{GSM8K} dataset, representing a standard implementation of translate-training approach. We also reproduce this model in our experiments, denoted as \textbf{MultiReason}.
    \item \textbf{MetaMath}, which is instruction-tuned with \textsc{MetaMathQA}~\cite{yu2023metamath}. It is currently the most powerful English instruction data for mathematical reasoning. We also reproduce this model in our experiments, denoted as \textbf{MonoReason}.
\end{itemize}
Among these baseline systems, most models are tuned with English data and only MathOctopus and MultiReason are tuned with multilingual data.

\noindent\paragraph{Evaluation Dataset}
To assess LLMs' performance on multilingual mathematical reasoning\footnote{In this paper, we evaluate LLMs' reasoning performance on ten languages: Bengali (Bn), Thai (Th), Swahili (Sw), Japanese (Ja), Chinese (Zh), German (De), French (Fr), Russian (Ru), Spanish (Es) and English (En).}, we employ the benchmark dataset \textsc{mGSM}~\cite{shi2022language}.
We also evaluate the robustness of LLMs using an out-of-domain test set \textsc{mSVAMP}~\cite{chen2023breaking}.
In our experiments, we report LLM's answer accuracy in a zero-shot and greedy decoding setting.
Specifically, we use evaluation scripts\footnote{\url{https://github.com/microsoft/MathOctopus}} provided by~\citet{chen2023breaking} and measure answer accuracy by comparing the last numerical number that appears in the LLM-generated response with the gold answer.

\begingroup
\renewcommand{\arraystretch}{1.2}
\begin{table*}[ht]
\centering
\footnotesize
\begin{tabular}{l|cccccccccc|c}
\hline
\hspace{1.2cm}\textbf{System (7B)} & \textbf{Bn} & \textbf{Th} & \textbf{Sw} & \textbf{Ja} & \textbf{Zh} & \textbf{De} & \textbf{Fr} & \textbf{Ru} & \textbf{Es} & \textbf{En} & \textbf{Avg.} \\
\hline
SFT$^\dagger$~\cite{touvron2023llama} & 11.5 & 18.2 & 17.2 & 31.6 & 35.2 & 39.0 & 39.1 & 39.1 & 39.2 & 38.8 & 30.9 \\ 
RFT$^\dagger$~\cite{yuan2023scaling} & 7.7  & 16.9 & 14.9 & 33.9 & 34.9 & 40.8 & 41.5 & 39.5 & 42.5 & 42.7 & 31.3 \\
MAmmoTH$^\dagger$~\cite{yue2023mammoth} & 4.3  & 6.3  & 4.2  & 26.7 & 26.8 & 39.6 & 39.9 & 33.7 & 42.9 & 45.1 & 26.3 \\
WizardMath$^\dagger$~\cite{luo2023wizardmath} & 16.1 & 17.0 & 10.3 & 37.9 & 36.3 & 39.2 & 37.7 & 37.4 & 44.8 & 48.5 & 32.5 \\
MathOctopus$^\dagger$~\cite{chen2023breaking} & 31.8 & 39.3 & 43.4 & 41.1 & 42.6 & 48.4 & 50.6 & 46.9 & 49.4 & 50.7 & 44.1 \\
MetaMath~\cite{yu2023metamath} & 14.2 & 17.8 & 16.5 & 53.2 & 53.1 & 61.4 & 60.7 & 58.9 & 61.2 & 65.5 & 46.3 \\ 
\hdashline
MultiReason & 27.6 & 36.5 & 42.4 & 40.9 & 43.2 & 44.3 & 46.7 & 42.3 & 45.5 & 48.0 & 41.3 \\
MonoReason & 15.0 & 17.1 & 15.4 & 51.9 & 54.4 & 60.9 & 62.2 & 59.3 & \textbf{63.3} & \textbf{65.5} & 46.2 \\
QAlign$\rightarrow$MonoReason (Ours)  & \textbf{41.7} & \textbf{47.7} & \textbf{54.8} & \textbf{58.0} & \textbf{55.7} & \textbf{62.8} & \textbf{63.2} & \textbf{61.1} & \textbf{63.3} & 65.3 & \textbf{57.2} \\
\hline
\hline
\hspace{1.2cm}\textbf{System (13B)} & \textbf{Bn} & \textbf{Th} & \textbf{Sw} & \textbf{Ja} & \textbf{Zh} & \textbf{De} & \textbf{Fr} & \textbf{Ru} & \textbf{Es} & \textbf{En} & \textbf{Avg.} \\
\hline
SFT$^\dagger$~\cite{touvron2023llama} & 13.9 & 23.4 & 19.8 & 41.8 & 43.3 & 46.2 & 47.8 & 47.8 & 46.1 & 50.9 & 38.1 \\ 
RFT$^\dagger$~\cite{yuan2023scaling}            & 12.2 & 24.8 & 19.4 & 42.4 & 42.3 & 45.1 & 45.2 & 46.5 & 45.6 & 47.1 & 37.1 \\
MAmmoTH$^\dagger$~\cite{yue2023mammoth}         & 5.0  & 13.7 & 12.9 & 42.2 & 47.7 & 52.3 & 53.8 & 50.7 & 53.9 & 53.4 & 38.6 \\
WizardMath$^\dagger$~\cite{luo2023wizardmath}   & 13.7 & 16.3 & 12.5 & 29.5 & 37.0 & 48.7 & 49.4 & 43.8 & 49.4 & 56.3 & 35.7 \\
MathOctopus$^\dagger$~\cite{chen2023breaking} & 35.2 & 41.2 & 46.8 & 39.2 & 52.0 & 47.2 & 48.0 & 45.6 & 53.2 & 56.4 & 46.5 \\
MetaMath~\cite{yu2023metamath} & 14.6 & 15.7 & 17.4 & 57.0 & 56.6 & 67.3 & 64.7 & 63.7 & 65.9 & 67.7 & 49.1 \\ 
\hdashline
MultiReason & 35.0 & 41.3 & 44.6 & 49.9 & 48.1 & 53.3 & 53.2 & 51.6 & 52.5 & 54.5 & 48.4 \\
MonoReason & 20.6 & 20.5 & 19.1 & 57.0 & 58.8 & 68.4 & \textbf{68.1} & \textbf{67.5} & \textbf{68.9} & \textbf{68.9} & 51.8 \\
QAlign$\rightarrow$MonoReason (Ours)   & \textbf{49.2} & \textbf{55.5} & \textbf{55.2} & \textbf{64.3} & \textbf{63.8} & \textbf{69.5} & \textbf{68.1} & 66.4 & 66.4 & 67.6 & \textbf{62.6} \\
\hline
\end{tabular}
\caption{Results on \textsc{mSVAMP} dataset. ``Avg.'' represents the average multilingual performance and bold text denotes the highest score among systems of the same size. The dagger symbol denotes that the results for these models are taken from the published results of~\citet{chen2023breaking}.}
\label{tab:msvamp}
\end{table*}
\endgroup

\section{Main Results}
In this section, we report our experiment results and introduce our main findings.

\subsection{Monolingual Supervision Setting}
\noindent\paragraph{Question alignment stage enables LLM's proficiency in English to be transferred to non-English tasks.}
Experiment results on the \textsc{mGSM} dataset are presented in Table~\ref{tab:mgsm}.
We can see that LLMs trained with augmented English data (RFT, MAmmoTH, WizardMath, MetaMath and MonoReason) typically underperform on non-English tasks, despite showing improved performance in English compared to SFT model.
The multilingual MathOctopus outperforms existing open-source models in terms of multilingual performance.
However, as we have discussed, the translated dataset can be out-dated quickly and keeping translating cutting-edge English instuction can also be prohibitive due to the high translation cost. 

Unlike the translate-training approach, our framework can easily utilize the most advanced English instruction data, e.g., \textsc{MetaMathQA}.
With the question alignment stage (QAlign), we successfully transfer model's proficiency in English to non-English languages. 
On average, this leads to a 11.2\% increase in accuracy for the 7B model and a 13.2\% increase in accuracy for the 13B model.
These substantial improvements on non-English languages significantly reduce LLM's performance gap between non-English and English tasks, thereby demonstrating the effectiveness of our devised method.

\noindent\paragraph{After question alignment, our fine-tuned LLM surpasses the translate-training baseline by a large margin}
More importantly, we observe that after question alignment, our fine-tuned LLM surpasses the translate-training baseline (MathOctopus) by a large margin. 
By transferring the model's expertise in English to non-English scenarios, our approach outperforms MathOctopus by an average margin of +9.6\% for the 7B model and +11.3\% for the 13B model.
These results again demonstrate the superiority of our method\footnote{In Appendix~\ref{sec:multireason}, we also report the results of using question translation data for stage I training and multilingual instruction data for stage II training. This provides a more direct comparison (QAlign$\rightarrow$MultiReason vs. MultiReason), and the added question alignment stage also improves multilingual performance in this setting.}.

\noindent\paragraph{Our fine-tuned LLMs also exhibit better robustness on the out-of-domain test set}
Apart from evaluating on \textsc{mGSM}, we further assess the robustness of our LLMs on the out-of-domain test set \textsc{mSVAMP} (Table~\ref{tab:msvamp}).
The findings are generally consistent with those from \textsc{mGSM} dataset.
Notably, compared to the unaligned counterpart (MonoReason), our model (QAlign$\rightarrow$MonoReason) achieves significant improvement in average multilingual performance, with gains of 11.0\% for the 7B model and 10.8\% for the 13B model.
Our method outperforms the translate-training approach (MathOctopus) by an even larger margin here, showing increases of 13.1\% for the 7B model and 16.1\% for the 13B model, which shows its more generalized and robust performance.

\begin{figure*}
\centering
\includegraphics[width=0.95\textwidth]{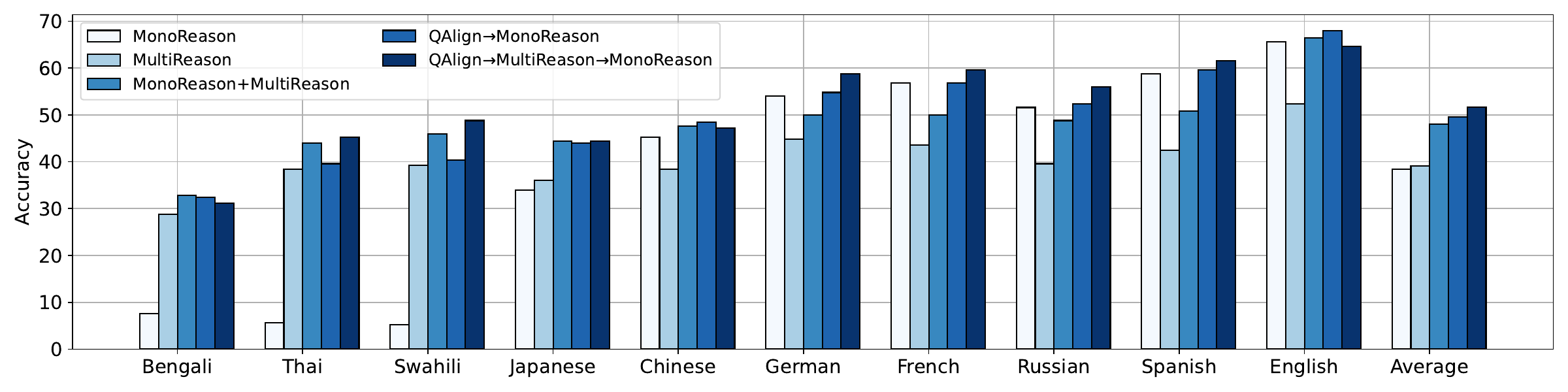}
\caption{Effects of tuning language-aligned LLM with mixed supervised data. Generally, incoporating multilingual supervised data into our framework can achieve a higher ceiling for average multilingual performance.}
\label{fig:mixed}
\end{figure*}
\subsection{Mixed Supervision Setting}
\noindent\paragraph{Incorporating multilingual supervised data into our framework can achieve a higher ceiling for multilingual performance}
Although our framework does not rely on the multilingual supervised data, we can utilize such data to attain a higher level of multilingual performance if a 
multilingual dataset is available.
In this mixed supervision setting, we first tune the stage I model (7B) with multilingual \textsc{GSM8KInstruct} and then tune it with English data \textsc{MetaMathQA}.
\begin{table}[ht]
\centering
\footnotesize
\resizebox{0.95\linewidth}{!}{
\begin{tabular}{ccccccc}
\toprule
\multirow{2}{*}{\textbf{Data}} & \multirow{2}{*}{\textbf{Direction}}& \multicolumn{2}{c}{\textbf{\textsc{mGSM}}} & \multicolumn{2}{c}{\textbf{\textsc{mSVAMP}}} \\
& & \textbf{Non-En} & \textbf{En} & \textbf{Non-En} & \textbf{En} \\
\midrule
\textit{Question} & X$\rightarrow$En & 47.6 & 68.0 & 56.5 & 65.3 \\ 
\textit{Question} & En$\rightarrow$X & 36.2 & 68.0 & 48.3 & 64.4 \\ 
\textit{Response} & X$\rightarrow$En    & 46.4 & 67.2 & 52.1 & 64.9 \\ 
\textit{Response} & En$\rightarrow$X    & 42.8 & 68.0 & 49.0 & 63.9 \\ 
\textit{Flores-101} & X$\rightarrow$En  & 36.3 & 68.0 & 46.8 & 65.4 \\ 
\bottomrule
\end{tabular}
}
\caption{Effects of using different translation training data for stage I training. ``X$\rightarrow$En'' and ``En$\rightarrow$X'' represents translating from non-English to English and translating English to non-English respectively. ``Non-En'' denotes LLM's average performance on non-English languages. Among these implementations, training LLM to translate non-English questions to English is the best one.}
\label{tab:ablation}
\end{table}
The experiment results on \textsc{mGSM} are depicted in Figure~\ref{fig:mixed}.
We find that incorporating additional multilingual supervision further leads to an average performance gain of 2.1\% on multilingual tasks.
Compared to the data mixing baseline (MonoReason+MultiReason), our approach demonstrate an average improvement of 3.6\%, with significant advantages in high-resource languages such as Spanish, Russian, German, and French.

\section{Analysis}
\subsection{Ablation study}
\noindent\paragraph{Impact of using different translation training data}
During the question alignment stage, we implement the translation task by training LLMs on translating questions from non-English to English.
Now we present the ablation study to show the effects of alternative implementations (Table~\ref{tab:ablation}). while different implementations yield similar performance in English, their impact on non-English peformance varies significantly.
For instance, training LLMs on reverse translation tasks greatly degenerates non-English performance (\textit{Question}:En$\rightarrow$X, \textit{Response}:En$\rightarrow$X).
Training LLM on translating CoT responses from non-English to English (\textit{Response}:X$\rightarrow$En) also results in lower performance compared to our original implementation.
\begin{figure*}[ht]
    \centering
    \includegraphics[width=0.95\textwidth]{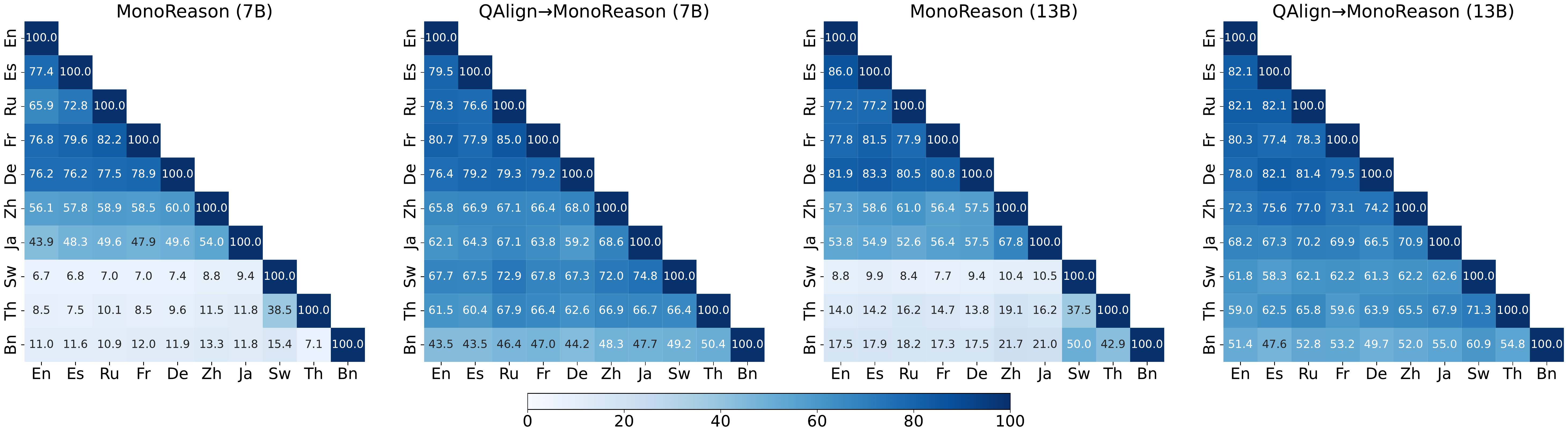}
   \caption{Comparing the prediction consistency of different systems. Darker blue denotes higher level of prediction consistency. Question alignment stage always brings improvement to the consistency of predicted answers.}
   \label{fig:consistency}
\end{figure*} 
\begin{table}[ht]
\centering
\resizebox{0.95\linewidth}{!}{
\begin{tabular}{lccccc}
\toprule
\multirow{2}{*}{\textbf{\hspace{1cm}Implementation}} & \multicolumn{2}{c}{\textbf{\textsc{mGSM}}} & \multicolumn{2}{c}{\textbf{\textsc{mSVAMP}}} \\
& \textbf{Non-En} & \textbf{En} & \textbf{Non-En} & \textbf{En} \\
\midrule
\large{\textit{our implementation}} & 47.6 & 68.0 & 56.5 & 65.3 \\ 
\large{\textit{$\hookrightarrow$ reversing training order}} & 2.0 & 2.8 & 2.0 & 2.0 \\ 
\large{\textit{$\hookrightarrow$ single-stage training}} & 3.7 & 68.0 & 2.6 & 65.2 \\ 
\bottomrule
\end{tabular}
}
\caption{Effects of reversing training order and performing single-stage multi-task training. Among these implementations, our original implementation, i.e., performing question alignment at first and then perform response alignment, is the best one.}
\label{tab:order}
\end{table} 
We suggest that this is because noises in the translated CoT responses compromise the data quality.
Training the LLM with translation data from commonly-used corpora, such as \textsc{Flores}\footnote{In this ablation study, we take the translation data in the development and test set of \textsc{Flores-101} dataset~\cite{goyal2022flores} for fine-tuning.}, does not work as well, indicating that the domain of the translation data is another crucial factor in establishing language alignment.

\noindent\paragraph{Impact of manipulating training order}
We also conduct the ablation study to demonstrate the significance of the training sequence within our proposed framework.
As shown in Table~\ref{tab:order}, reversing the order of the two training stages results in the LLM performing poorly in both English and non-English languages.
We observed that an LLM fine-tuned in this manner tends to repeat the question in English when presented with questions in various languages.

When we merge the training datasets from both stages and perform a single-stage, multi-task training, there is a significant drop in non-English performance as well. 
Although capable of responding to questions in English, the fine-tuned LLM is prone to translating the given non-English questions rather than answering them.
These analysis results demonstrate that our design of two-step training framework is non-trivial.

\begin{table}[ht]
\centering
\footnotesize
\resizebox{0.95\linewidth}{!}{
\begin{tabular}{cccccc}
\toprule
\multirow{2}{*}{\textbf{Method}} & \multicolumn{2}{c}{\textbf{\textsc{mGSM}}} & \multicolumn{2}{c}{\textbf{\textsc{mSVAMP}}} \\
& \textbf{Non-En} & \textbf{En} & \textbf{Non-En} & \textbf{En} \\
\midrule
\multicolumn{5}{c}{\textbf{MonoReason} (7B)} \\
Direct Inference & 35.4 & 65.5 & 47.6 & 68.9 \\ 
Translate-test & 30.8 & -    & 42.3 & -  \\ 
\midrule 
\multicolumn{5}{c}{\textbf{QAlign$\rightarrow$MonoReason} (7B)} \\
Direct Inference & 47.6 & 68.0 & 56.5 & 65.3 \\ 
Translate-test   & 46.6 & -    & 56.6 & -    \\ 
\bottomrule
\end{tabular}
}
\caption{Comparison between direct inference and translate-test inference.}
\label{tab:transtest}
\end{table}
\subsection{Prediction Consistency}
Another advantage of establishing question alignment is the improvement it brings to the consistency\footnote{Supposing the set of correct predictions in two languages is $U$ and $V$ respectively, we compute the consistency score as $\frac{|U\cap V|}{|U|}$.} of predicted answers against multilingual queries.
This means a higher degree of agreement in answers to the same question posed in different languages.
Figure~\ref{fig:consistency} displays the quantified results.
In contrast to their unaligned counterparts (MonoReason), our alignment-enhanced LLM (QAlign$\rightarrow$MonoReason) usually demonstrate higher answer consistency.
This improvement is particularly notable for distant languages, such as Bengali, Thai, Swahili, Japanese, and Chinese.
This results can serve as another strong evidence of our successful transfer of LLM's proficiency in English to non-English languages.
Appendix~\ref{sec:consistency} presents some cases to further illustrate the advantages of achieving higher multilingual consistency.

\subsection{Question Alignment vs Translate-Test}
\begin{table}[ht]
\centering
\footnotesize
\resizebox{0.95\linewidth}{!}{
\begin{tabular}{cccccc}
\toprule
\multirow{2}{*}{\textbf{Supervision}} & \multirow{2}{*}{\textbf{QAlign}} & \multicolumn{2}{c}{\textbf{\textsc{mGSM}}} & \multicolumn{2}{c}{\textbf{\textsc{mSVAMP}}} \\
  &  & \textbf{Non-En} & \textbf{En} & \textbf{Non-En} & \textbf{En} \\
\midrule
\textsc{GSM8K} & \XSolidBrush & 18.8 & 43.6 & 33.6 & 47.2 \\ 
\textsc{GSM8K} & \Checkmark   & 26.3 & 41.6 & 36.8 & 47.0 \\ 
\midrule 
\textsc{MetaMathQA} & \XSolidBrush & 35.4 & 65.6 & 44.4 & 65.3 \\
\textsc{MetaMathQA} & \Checkmark   & 47.6 & 68.0 & 56.5 & 65.3 \\ 
\bottomrule
\end{tabular}
}
\caption{Effects of tuning the stage I model (7B) with different English instruction data.}
\label{tab:instruct}
\end{table}
In our training framework, we implicitly endow the LLM with a bias that associates non-English questions with their English equivalents, sharing similar philosophy with translate-test prompting approach.
Thus we discuss the difference between these two approaches here.
Experiment results are reported in Table~\ref{tab:transtest}.
For the MonoReason model, the translate-test approach does not yield any improvement, suggesting that this approach may not be universally applicable solution for open-source LLMs.
For our alignment-enhanced QAlign$\rightarrow$MonoReason model, direct inference and translate-test prompting achieves similar performance.
But considering our approach does not rely on explicitly translating the questions during inference, it will have a more efficient inference process.

\subsection{Effects of tuning LLM with different English instruction data}
To demonstrate the universal effectiveness of question alignment, we also employ English \textsc{GSM8K} dataset as monolingual supervison and show the results in Table~\ref{tab:instruct}.
Under different English instruction data, the incorporation of a question alignment stage always boost LLM's non-English performance.
These results also highlight the importance of using advanced English instruction data, because achieving better performance in English usually means an improved non-English performance with the help of inner language alignment. 

\section{Conclusion}
In this paper, we introduce a novel question alignment method to empower LLMs on multilingual mathematical reasoning tasks without requiring multilingual instruction data. 
Experiment results on \textsc{mGSM} and \textsc{mSVAMP} benchmarks show that our proposed question alignment stage brings an average improvement of up to 13.2\% in multilingual performance. 
Our alignment-enhanced LLM outperforms the unaligned counterpart and the translate-training baseline by a large margin and shows a more robust performance.
Generally, our devised method successfully narrows the gap between LLM's performance between English and non-English, showing a new possibility to unlock LLM's capabilities to solve multilingual tasks.

\section*{Limitation}
Below we discuss potential limitations of our work:
\begin{itemize}[itemsep=1pt]
    \item Chain-of-Thought in English: When receiving non-English questions, our language-aligned LLM typically produces an English CoT before giving the final numerical answer. While the language used for the CoT is not explicitly specified as a requirement for the multilingual mathematical reasoning task, providing a CoT consistent with the query's language could enhance the model's utility.
    \item Scale of the pre-trained LLM: Our experiment is constrained by available computational resources, leading us to utilize the LLaMA2-7B and LLaMA2-13B models. Should resources allow in the future, we aim to broaden our research to include larger-scale models, such as LLaMA2-70B.
\end{itemize}

\section*{Acknowledgement}
We would like to thank the anonymous reviewers for their insightful comments. 
Shujian Huang is the corresponding author. 
This work is supported by National Science Foundation of China (No. 62376116, 62176120), the Liaoning Provincial Research Foundation for Basic Research (No. 2022-KF-26-02), research project of Nanjing University-China Mobile Joint Institute.
This project has also received funding from UK Research and Innovation (UKRI) under the UK government's Horizon Europe funding guarantee (UTTER grant numbers 10039436).
Wenhao Zhu is also supported by China Scholarship Council (No.202306190172).

\normalem
\bibliography{custom}

\appendix

\clearpage
\appendix
\section{Analyzing the Quality of the Translated Dataset}
\label{sec:quality}
In the work of~\citep{chen2023breaking}, the authors employ ChatGPT to translate \textsc{GSM8K} into several non-English languages, resulting in the creation of the multilingual dataset \textsc{GSM8KInstruct}.
Below we analyze the translation quality of this dataset and highlight the challenges associated with translating complex CoT responses.
We evaluate the translation quality of both questions and responses in a reference-free condition with COMETKiwi\footnote{Specifically, we employ \textit{wmt22-cometkiwi-da} as the evaluation model: \url{https://huggingface.co/Unbabel/wmt22-cometkiwi-da}.}~\cite{rei2022cometkiwi}. 
The evaluation results in Table~\ref{tab:comet} show that the quality of the translated responses is significantly inferior to that of the translated questions. 
This gap demonstrates the difficulties inherent in translating CoT content. 

Table~\ref{tab:error} provides some examples of typical translation errors.
Based on this analysis, we suggest that constructing a multilingual CoT dataset through a translation engine is fraught with errors and cannot ensure the quality of the dataset.
In constrast, our devised framework provides a more effective and efficient solution, which does not require translated multilingual CoT.

\begingroup
\renewcommand{\arraystretch}{1.2}
\begin{table*}[ht]
\centering
\footnotesize
\begin{tabular}{cccccccccc}
\toprule
\multirow{2}{*}{Analyzed Data} & \multicolumn{9}{c}{COMETKiwi (En-X)} \\
\cline{2-10}
      & Bn    & Th    & Sw    & Ja    & Zh    & De    & Fr    & Ru    & Es    \\
\midrule
\textit{Question Translation} & 82.22 & 79.61 & 82.60 & 86.64 & 82.95 & 83.56 & 82.29 & 84.53 & 85.59 \\
\textit{Response Translation} & 79.92 & 76.99 & 76.84 & 83.34 & 79.30 & 78.34 & 79.98 & 79.85 & 79.83 \\
\bottomrule
\end{tabular}
\caption{Evaluation results of the translation quality of \textsc{GSM8KInstruct} dataset.}
\label{tab:comet}
\end{table*}
\begingroup

\begingroup
\renewcommand{\arraystretch}{1.2}
\begin{table*}[ht]
\centering
\scalebox{0.9}
{
\begin{tabular}{p{18cm}}
\toprule
\textbf{Example I}: missing certain reasoning step in the translated response \\
\midrule
$[\textbf{\textrm{English CoT Response}}]$\newline
Half of the wallet's price is \$99 / 2 = \$50.\newline
Betty's grandparents gave her \$15 * 2 = \$30.\newline
In total, Betty has \$50 + \$15 + \$30 = \$95.\newline
So she still needs \$100 – \$95 = \$5 which her parents plan to give her for her birthday. \\
\midrule
$[\textbf{\textrm{Translated Chinese CoT Response}}]$\newline
\kaiti{钱包价格的一半是100美元 / 2 = 50美元。} \newline
\textcolor{gray}{(Half the wallet price is \$100/2=\$50.)} \newline
\kaiti{贝蒂的祖父母给了她15美元 * 2 = 30美元。} \newline
\textcolor{gray}{(Betty's grandparents gave her \$15*2=\$30.)} \newline
\kaiti{这意味着，贝蒂还需要50美元 - 30美元 - 15美元 = 5美元。} \newline
\textcolor{gray}{(This means that Betty still needs \$50 - \$30 - \$15 = \$5.)} \\
\midrule
\textbf{Example II}: distinct reasoning paths exist between responses in English and the translated version \\
\midrule
$[\textbf{\textrm{English CoT Response}}]$\newline
The total time he takes to walk to work 3 times a week is 2*3 = 6 hours\newline
The total time he takes to walk to and from work 3 times a week is 6+6 = 12 hours\newline
The total time he takes to bike to work 2 times a week is 2*1 = 2 hours\newline
The total time he takes to bike to and from work 2 times a week is 2+2 = 4 hours\newline
The total time he takes to go to and from work in a week is 12+4 = 16 hours \\
$[\textbf{\textrm{Translated Chinese CoT Response}}]$\newline
\kaiti{Roque每周走路去上班需要2*3=6小时。} \newline 
\textcolor{gray}{(It takes Roque 2*3=6 hours to walk to work every week.)} \newline
\kaiti{Roque每周骑自行车去上班需要1*2=2小时。} \newline
\textcolor{gray}{(It takes Roque 1*2=2 hours to ride a bicycle to work every week.)}\newline
\kaiti{他每周去上班的总时间是6+2=8小时。} \newline 
\textcolor{gray}{(The total time he goes to work every week is 6+2=8 hours.)} \newline
\kaiti{他每周回家的总时间也是8小时。} \newline 
\textcolor{gray}{(The total time he spends home every week is also 8 hours.)} \newline
\kaiti{因此，Roque每周去上班和回家总共需要8+8=16小时。} \newline 
\textcolor{gray}{(Therefore, it takes Roque a total of 8+8=16 hours per week to get to work and back home.)} \\
\bottomrule
\end{tabular}
}
\caption{Examples of failing to accurately translating English CoT response. For illustration, the gray text is the English meaning of the corresponding Chinese sentence.}
\label{tab:error}
\end{table*}
\endgroup

\section{Experiment Results of Using Multilingual Instruction Data for Response Alignment}
\label{sec:multireason}
To more comprehensively illustrate the benefit of the question alignment approach, we use question translation data for stage I training and multilingual instruction data GSM8KInstruct for stage II training (QAlign$\rightarrow$MultiReason). 
We can see that the added question alignment stage also brings improvement on multilingual performance in this setting. 

\begingroup
\renewcommand{\arraystretch}{1.2}
\begin{table*}[ht]
\centering
\footnotesize
\begin{tabular}{l|cccccccccc|c}
\hline
\hspace{0.6cm}\textbf{System (7B)} & \textbf{Bn} & \textbf{Th} & \textbf{Sw} & \textbf{Ja} & \textbf{Zh} & \textbf{De} & \textbf{Fr} & \textbf{Ru} & \textbf{Es} & \textbf{En} & \textbf{Avg.} \\
\hline
MultiReason & 26.8 & 36.0 & 36.8 & 33.2 & 42.4 & 42.8 & \textbf{40.8} & \textbf{42.4} & 42.8 & 47.2 & 39.1 \\
QAlign$\rightarrow$MultiReason & \textbf{31.6} & \textbf{36.4} & \textbf{38.8} & \textbf{38.0} & \textbf{43.6} & \textbf{45.2} & \textbf{40.8} & 38.4 & \textbf{46.8} & \textbf{49.2} & \textbf{40.9} \\
\hline
\end{tabular}
\caption{Results on \textsc{mGSM} dataset. ``Avg.'' represents the average multilingual performance and bold text denotes the highest score among systems of the same size.}
\label{tab:msvamp}
\end{table*}
\endgroup

\section{Benefits of Achieving Higher Multilingual Consistency}
\label{sec:consistency}
Table~\ref{tab:case} presents some examples to illustrate the advantages of achieving higher multilingual consistency.
In this case, both MonoReason model and QAlign$\rightarrow$MonoReason model predict correctly on the English question.
However, when presented with the same question expressed in Chinese, our LLM generates a similar reasoning path and still give the right answer, but the unaligned model fails.

\begin{table*}[ht]
\centering
\scalebox{0.9}
{
\begin{tabular}{p{14cm}p{2cm}<{\centering}}
\toprule
\hspace{5.7cm} \textbf{Question (English)} & \textbf{Gold Answer} \\
\midrule
There are 27 unicorns left in the world.  One third of them are in the Scottish Highlands.  Two thirds of the Scottish unicorns are female.  How many female Scottish unicorns are there? & 6 \\
\midrule 
\hspace{5.2cm} \textbf{Chain-of-Thought Response} & \textbf{Prediction} \\
\midrule
$[\textbf{\textrm{MonoReason}}]$ \newline One third of the unicorns are in the Scottish Highlands, so there are 27/3 = 9 unicorns in the Highlands.\newline Two thirds of the Scottish unicorns are female, so there are 9 * 2/3 = 6 female Scottish unicorns.\newline The answer is: 6 & \hspace{0.3cm} 6 \large\textcolor{green}{\ding{51}} \\ 
\midrule 
$[\textbf{\textrm{QAlign$\rightarrow$MonoReason}}]$ \newline One third of the unicorns are in the Scottish Highlands, so there are 27/3 = 9 unicorns in the Highlands.\newline Two thirds of the unicorns in the Highlands are female, so there are 9 * 2/3 = 6 female unicorns in the Highlands.\newline Therefore, there are 6 female Scottish unicorns.\newline The answer is: 6 & \hspace{0.3cm} 6 \large\textcolor{green}{\ding{51}} \\
\midrule
\hspace{5.7cm} \textbf{Question (Chinese)} & \textbf{Gold Answer} \\
\midrule
\kaiti{世界上剩下 27 只独角兽，其中三分之一在苏格兰高地。\newline 苏格兰高地的独角兽的三分之二是雌性。苏格兰雌性独角兽有多少只？} & 6 \\
\midrule
\hspace{5.2cm} \textbf{Chain-of-Thought Response} & \textbf{Prediction} \\
\midrule
$[\textbf{\textrm{MonoReason}}]$ \newline If three-fourths of the remaining 27 one-horned animals are in the Scottish Highlands, then there are 27 * 3/4 = 21 one-horned animals in the Scottish Highlands. \newline If two-thirds of the one-horned animals in the Scottish Highlands are female, then there are 21 * 2/3 = 14 female one-horned animals in the Scottish Highlands. \newline Therefore, there are 14 female one-horned animals in the Scottish Highlands. \newline The answer is: 14 & \hspace{0.3cm} 14 \large\textcolor{red}{\ding{55}} \\
\midrule
$[\textbf{\textrm{QAlign$\rightarrow$MonoReason}}]$ \newline If there are 27 remaining unicorns in the world, and one-third of them are in the Scottish Highlands, then there are 27/3 = 9 unicorns in the Scottish Highlands.\newline If two-thirds of the unicorns in the Scottish Highlands are female, then there are 9 * 2/3 = 6 female unicorns in the Scottish Highlands.\newline Therefore, there are 6 female unicorns in the Scottish Highlands. \newline The answer is: 6 & \hspace{0.3cm} 6 
\large\textcolor{green}{\ding{51}} \\
\bottomrule
\end{tabular}
}
\caption{With this case we illustrate the advantages of achieving higher prediction consistency. Given the same question expressed in different languages, our alignment-enhanced model (QAlign$\rightarrow$MonoReason) can generate a similar reasoning path and give the right answer consistently.}
\label{tab:case}
\end{table*}

\section{Used Scientific Artifacts}
Below lists scientific artifacts that are used in our work. For the sake of ethic, our use of these artifacts is consistent with their intended use.
\begin{itemize} [itemsep=1pt]
    \item \textit{Stanford Alpaca (Apache-2.0 license)}, a project that aims to build and share an instruction-following LLaMA model. 
    \item \textit{Transformers (Apache-2.0 license)}, a framework that provides thousands of pretrained models to perform tasks on different modalities such as text, vision, and audio.
\end{itemize}

\end{document}